\documentclass[smallcondensed,final]{svjour3}
\usepackage{graphicx}
\usepackage{epstopdf}

\newcommand{\bc}{\begin{equation*}\begin{array}{l}}
\newcommand{\ec}{\end{array}\end{equation*}}
\newcommand{\com}[1]{}

\begin{document}

\title{Constructing Runge-Kutta Methods with the Use of Artificial Neural Networks
\footnote{The final publication is available at http://link.springer.com/article/10.1007\%2Fs00521-013-1476-x}}

\author{Angelos A. Anastassi}

\institute{A.A. Anastassi \at
    Department of Informatics,\\
    University of Piraeus,\\
    Karaoli \& Dimitriou 80,\\
    18534 Piraeus, GREECE\\
    \email{ang.anastassi@gmail.com}
}

\date{}

\maketitle

\begin{abstract}
A methodology that can generate the optimal coefficients of a numerical method with the use of an artificial neural network is presented in this work. The network can be designed to produce a finite difference algorithm that solves a specific system of ordinary differential equations numerically. The case we are examining here concerns an explicit two-stage Runge-Kutta method for the numerical solution of the two-body problem. Following the implementation of the network, the latter is trained to obtain the optimal values for the coefficients of the Runge-Kutta method. The comparison of the new method to others that are well known in the literature proves its efficiency and demonstrates the capability of the network to provide efficient algorithms for specific problems.
\end{abstract}


\section{Introduction}
\label{Intro}

The literature that involves solving ordinary or partial differential equations with the use of artificial neural networks is quite limited, but has grown significantly in the past decade. To name a few, Dissanayake et al. used a "universal approximator" (see \cite{cybenko}\cite{hornik}) to transform the numerical problem of solving partial differential equations to an unconstrained minimization problem of an objective function \cite{dissanayake}. Meade et al. demonstrated feedforward neural networks that could approximate linear and nonlinear ordinary differential equations \cite{meade}\cite{meade2}. Puffer et al. constructed cellular neural networks that were able to approximate the solution of various partial differential equations \cite{puffer}. Lagaris et al. introduced a method for the solution of initial and boundary value problems, that consists of an invariable part, which satisfies by construction the initial/boundary conditions and an artificial neural network, which is trained to satisfy the differential equation \cite{lagaris}. S. He et al. used feedforward neural networks to solve a special class of linear first-order partial differential equations \cite{he}. The radial basis function (RBF) network architecture \cite{alexandridis} has also been used for the solution of differential equations in \cite{jianyu}, where Jianyu et al. demonstrated a method for solving linear ordinary differential equations based on multiquadric RBF networks. Ramuhalli et al. proposed an artificial neural network with an embedded finite-element model, for the solution of partial differential equations \cite{ramuhalli}. Tsoulos et al. demonstrated a hybrid method for the solution of ordinary and partial differential equations, which employed neural networks, based on the use of grammatical evolution, periodically enhanced using a local optimization procedure \cite{tsoulos}.

In all the aforementioned cases, the neural networks functioned as direct solvers of differential equations. In this work, however, we use the constructed neural network, not as a direct solver, but as a means to generate proper Runge-Kutta coefficients. In this aspect, there has been little relevant published material. For instance in \cite{tsitouras}, Tsitouras constructs a multilayer feedforward neural network that uses input data associated with an initial value problem and trains the network to produce the solution of this problem as output data, thus generating (by construction of the network) coefficients for a predefined number of Runge-Kutta stages.

In this work we construct an artificial neural network that can generate the coefficients of two-stage Runge-Kutta methods. We consider the two-body problem, which is a typical case of an initial value problem where Runge-Kutta methods apply, and therefore the resulting method is specialized in solving it. The comparison shows that the new method is more efficient than the classical methods and thus proves the capability of the constructed neural network to create new Runge-Kutta methods.

The structure is as follows: in Sections \ref{Sec_RK} and \ref{Sec_ANN} we present the basic theory for Runge-Kutta methods and Artificial Neural Networks respectively. In Section \ref{Sec_Implementation} we present the implementation of the neural network that applies to a two-stage Runge-Kutta method which solves the two-body problem numerically, while in Section \ref{Sec_Derivation} the derivation of the method is provided. In Section \ref{Sec_Results} we demonstrate the final results along with the comparison of the new method to other well-known methods and finally in Section \ref{Sec_Conclusions} we reach some conclusions about this work.

\section{Runge-Kutta methods}
\label{Sec_RK}

We consider a two-stage Runge-Kutta method to solve the first order initial value problem

\begin{equation}
\label{secondorder}
 {\bf v}'(t) = f(t,{\bf v}) \, \, \, \mbox{and} \, \, \, {\bf v}(t_0)={\bf v_0}
\end{equation}

\noindent At each step of the integration interval, we approximate the solution of the initial value problem at $t_{n+1}$, where $t_{n+1} = t_n + h$. For the two-stage Runge-Kutta method, the approximate solution \({\bf v_{n+1}}\) is given by

\begin{equation}
\label{RK}
 {\bf v_{n+1}} = {\bf v_n} + h \, (b_1 \, {\bf k_1} + b_2 \, {\bf k_2})
\end{equation}

\noindent where

\begin{equation}
 {\bf k_1} = f(t_n , {\bf v_n})
\end{equation}
\begin{equation}
 {\bf k_2} = f(t_n + c_2 \, h, {\bf v_n} + h \, a_{21} \, {\bf k_1})
\end{equation}

\noindent The coefficients for this set of methods can be presented by the Butcher tableau given below

\begin{table}[hb]
\begin{tabular}{l|l l}
 $c_2$ & $a_{21}$ & \\
 \hline
 & $b_1$ & $b_2$
\end{tabular}
\end{table}

\noindent An explicit two-stage Runge-Kutta method can be of second algebraic order at most (see \cite{butcher}). In order for that to hold, the following conditions must be satisfied

\begin{equation}
\label{alg1}
b_1 + b_2 = 1
\end{equation}
\begin{equation}
\label{alg2}
b_2 \, c_2 = \frac{1}{2}
\end{equation}

The extra condition that needs to be satisfied in order for the method to be consistent is
\begin{equation}
\label{cons}
a_{21} = c_2
\end{equation}

\section{Artificial Neural Networks}
\label{Sec_ANN}

An {\bf artificial neural network (ANN)} is a network of interconnected artificial processing elements (called {\bf neurons}) that co-operate with each other in order to solve specific problems. ANNs are inspired by the structure and functional aspects of biological nervous systems and therefore present a resemblance. Haykin in \cite{haykin} defines an artificial neural network as "a massively parallel distributed processor that has a natural propensity for storing experiential knowledge and making it available for use". ANNs, similarly to brains, acquire knowledge through a learning process, which is called {\bf training}. That knowledge is stored in the form of {\bf synaptic weights}, whose values express the strength of the connection between two neurons.

There are many types of artificial neural networks, depending on the structure and the means of training. An ANN, in its simplest form, consists of a single neuron, in what we call the {\bf Perceptron} model. The Perceptron is connected with a number of {\bf inputs}: $x_1$, $x_2$, ..., $x_n$. A weight corresponds to each of the neuron's connections with the inputs and expresses the specific input's significance to the calculation of the Perceptron's output. Therefore, there is a number of weights equal to the number of inputs, with $w_i$ being the weight that corresponds to input $x_i$, and $w_i x_i$ being the combined weighted input. The neuron is also connected with a weight, which is not connected with any input and is called {\bf bias} (symbolized by $b$). The weighted input data are essentially being mapped to an {\bf output} value, through the {\bf transfer function} of the neuron. The transfer function consists of the {\bf net input function} and the {\bf activation function}. The first function undertakes the task of combining the various weighted inputs into a scalar value. Typically, the summation function is used for this purpose, thus the {\bf net input} in that case is described by the following expression

\begin{figure}
\includegraphics[width = \textwidth]{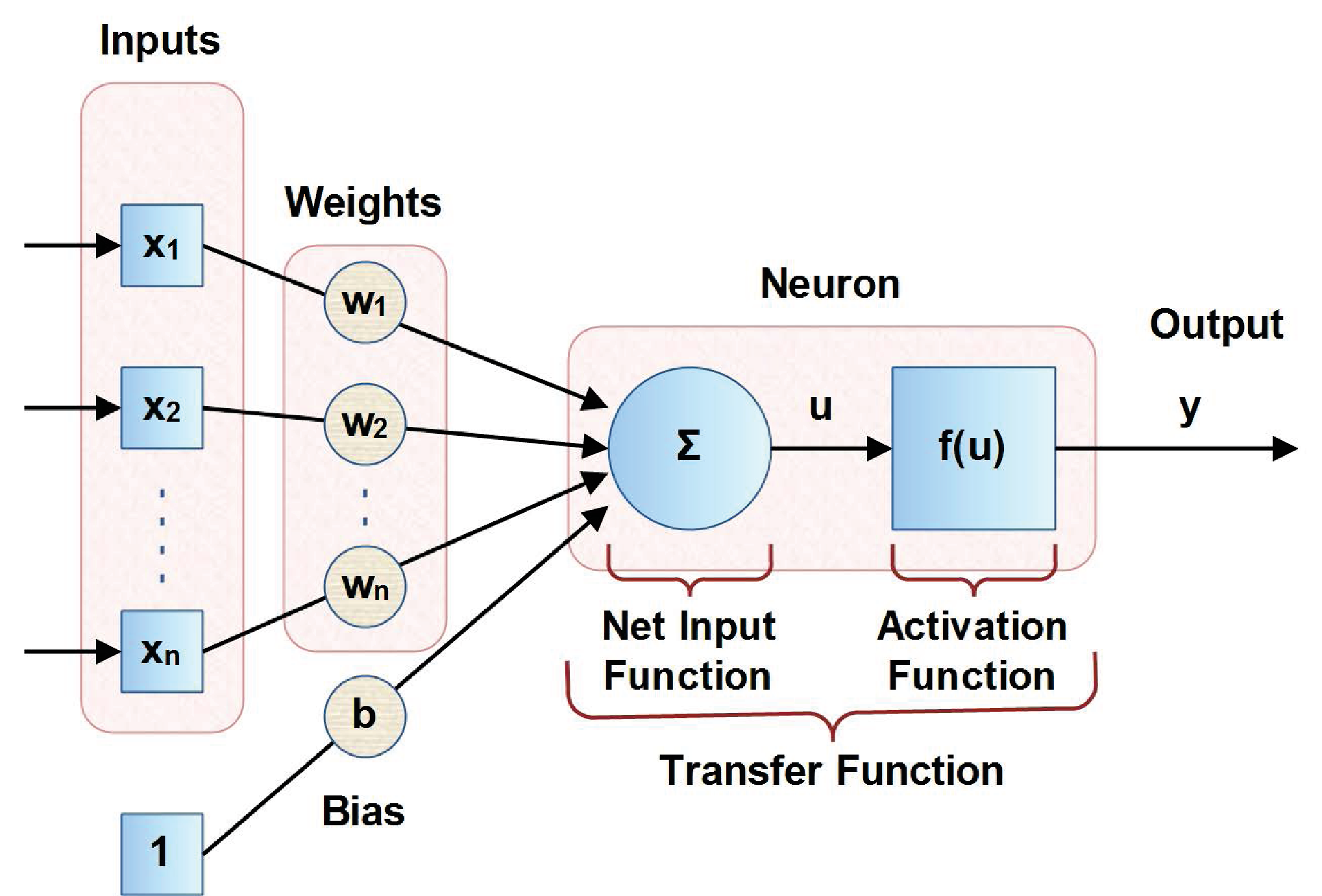}
\caption{An artificial neuron (Perceptron)}
\end{figure}

\begin{equation}
u = \displaystyle\sum\limits_{i=1}^n w_i x_i + b
\end{equation}

For ease of use, the bias can be treated as any other weight and be represented as $w_0$, with $x_0 = 1$, so the previous expression becomes
\begin{equation}
u = \displaystyle\sum\limits_{i=0}^n w_i x_i
\end{equation}

The activation function receives the output $u$ of the net input function and produces its own output y (also known as the {\bf activation value} of the neuron), which is a scalar value as well. The general form of the activation function is
\begin{equation}
y = f(u)
\end{equation}

What is required of a neural network, such as the Perceptron, is the ability to learn. That means the ability to adjust its weights in order to be able to produce a specific set of {\bf output data} for a specific set of {\bf input data}. In the {\bf supervised} type of learning (which is used to train the Perceptron and other types of neural networks), besides the {\bf input data}, we provide the network with {\bf target data}, which are the data that the output should ideally converge to, with the given set of input. The Perceptron uses an estimation of the {\bf error} as a measure that expresses the divergence between the actual output and the target. In that sense, {\bf training} is essentially the learning process of adjusting the synaptic weights, in order to minimize the aforementioned error. To achieve this, a {\bf gradient descent} algorithm called {\bf delta rule} is employed, also known as the {\bf Least Mean Square (LMS) method} or the {\bf Widrow-Hoff rule}, developed by Widrow and Hoff in \cite{widrow-hoff}. The training process that uses the delta rule has the following steps:
\begin{enumerate}
\item Assign random values to the weights
\item Generate the output data for the set of the training input data
\item Calculate the error $E$, which is given by a norm of the differences between the target and the output data
\item Adjust the weights according to the following rule

\begin{equation}
\label{weight}
\Delta w_i = - \eta \frac{\partial E}{\partial w_i} = - \eta \frac{\partial E}{\partial u} \frac{\partial u}{\partial w_i} = \eta \delta x_i
\end{equation}

Where $\eta$ is a small positive constant, called {\bf learning step} or {\bf learning rate}, and $\delta$ ({\bf delta error}) is equal to the following expression

\begin{equation}
\label{delta}
\delta = -\frac{\partial E}{\partial u}
\end{equation}

\item Repeat from step 2, unless the error $E$ is less than a small predefined constant $\epsilon$ or the maximum number of iterations has been reached
\end{enumerate}

With each iteration of the learning method, the weights are adjusted in such manner as to reduce the error. Notice that the correction $\Delta w_i$ of the weights is proportional to the negative of $\frac{\partial E}{\partial w_i}$, since the desired goal is the minimization of the error.

Strictly speaking, the Perceptron is not technically a network, since it consists of a single neuron. An actual network and the most common form of ANNs is the {\bf multilayer Perceptron (MLP)}, which falls in the general category of {\bf feedforward neural networks (FFNNs)}. Feedforward are the neural networks that contain no feedback connections, i.e. the connections do not form a loop, but are instead all directed towards the output of the network. A multilayer Perceptron consists of various layers, namely the {\bf input layer}, the {\bf output layer}, and one or more {\bf hidden layers}. The hidden layers are not visible externally of the network, hence the "hidden" property. Each of the layers consists of one or more neurons, except for the input layer, whose input nodes simply function as an entry point for the input data. Each layer is connected to the previous and the next layer, thus providing a pathway for the data to travel throughout the network. When a layer is connected to another, each neuron of the first layer is connected to every neuron of the second layer. In this way, the output of one layer becomes the input for the next one. Therefore, to calculate the output of the MLP, the data (either input data or activation values) are being propagated forward through the neural network.

\begin{figure}
\includegraphics[width = \textwidth]{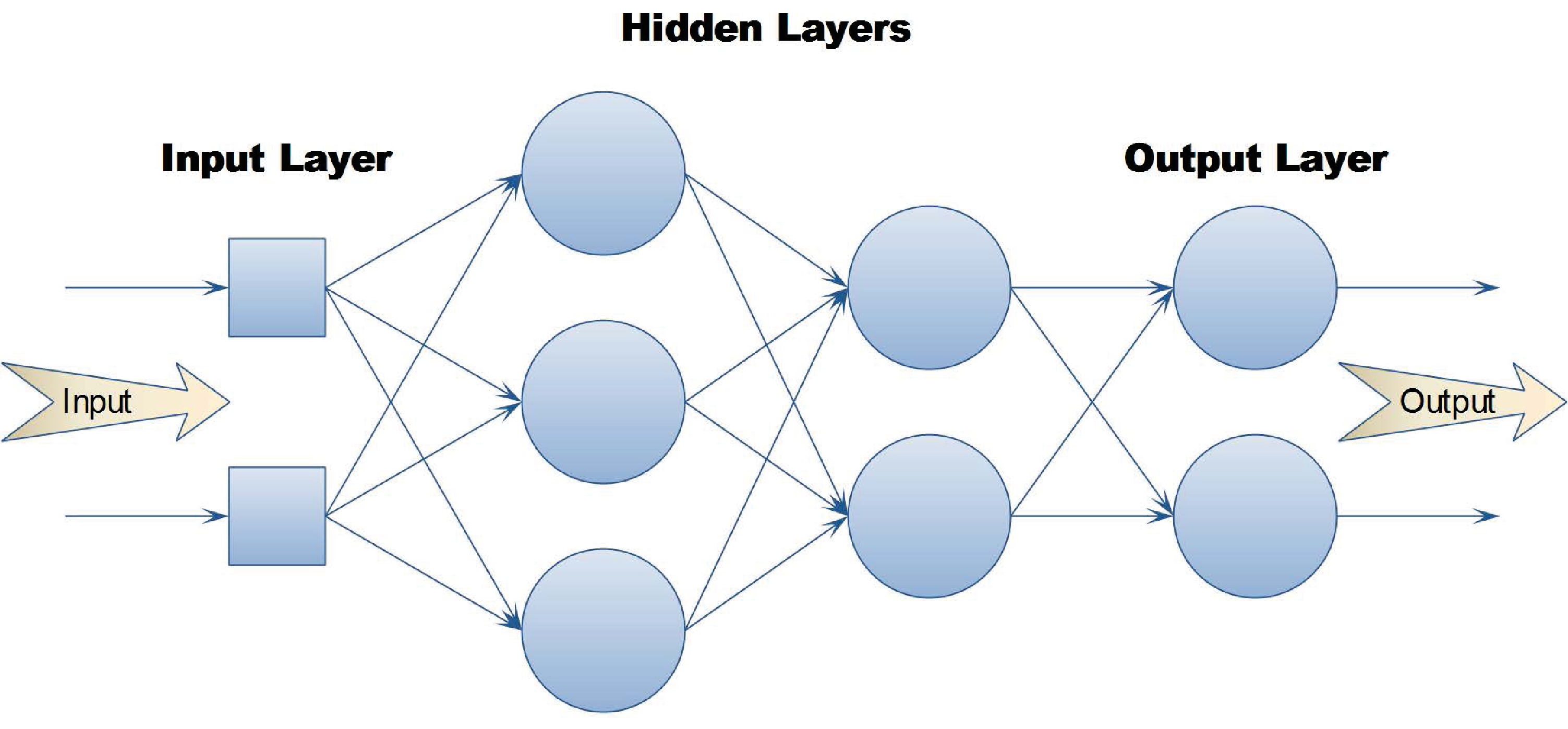}
\caption{A typical multilayer Perceptron}
\end{figure}

To train the MLP, the {\bf backpropagation} method is used. Backpropagation is a generalization of the delta rule that is used to train the Perceptron. The method, essentially, functions in the same way for each neuron of the MLP, as simple delta rule does for the Perceptron. The difference lies in the fact that, since the learning procedure requires the calculation of the $\delta$ error, the neurons that reside inside the hidden layers must be provided with the $\delta$ errors of the neurons of the next layer. For this reason, the $\delta$ errors are first calculated for the neurons of the output layer and are propagated backwards (hence the backpropagation term), all across the network. The procedure is the following:
\begin{enumerate}
\item Assign random values to the weights
\item Generate the activation values of each neuron, starting from the first hidden layer and continue until the output layer
\item Calculate the error $E$, which is given by a norm of the differences between the target and the output data
\item Calculate the $\delta$ errors of each neuron of the output layer, according to (\ref{delta})
\item Calculate the $\delta$ errors of each neuron of the previous layer, according to

\begin{equation}
\delta = f'(u) \displaystyle\sum\limits_{i=1}^M w_i \delta_i
\end{equation}

Where $f(u)$ is the activation function of the neuron, $M$ is the number of the neurons in the next layer, and $\delta_i$ is the $\delta$ error of the $i$th neuron of the next layer
\item Repeat step 5 for the previous layer, until all $\delta$ errors for all the hidden layers have been calculated
\item Adjust the weights according to (\ref{weight})
\item Repeat from step 2, unless the error $E$ is less than a small predefined constant $\epsilon$ or the maximum number of iterations has been reached
\end{enumerate}

\section{Implementation}
\label{Sec_Implementation}

We introduce a feedforward neural network that can generate the coefficients of two-stage Runge-Kutta methods, specifically designed to apply to the needs of the two-body problem. The two-body problem is described by the following system of equations
\begin{equation}
x'' = - \frac{x}{r^3}, \; y'' = - \frac{y}{r^3}, \; r = \sqrt{x^2 + y^2}
\end{equation}

\noindent whose analytical solution is given by

\begin{equation}
x = \cos{t}, \quad y = \sin{t}
\end{equation}

Since a Runge-Kutta method can only solve a system of first order ordinary differential equations, ${\bf v}$ according to the notation of equation (\ref{secondorder}), can be given by
\begin{equation}
 {\bf v} = [v_1, v_2, v_3, v_4] = \left[x, \, y, \, x', \, y'\right]
\end{equation}

\noindent and

\begin{equation}
 f(t,{\bf v}) = {\bf v}'(t) = \left[v_1', v_2', v_3', v_4'\right] = \left[v_3, \, v_4,  \, - \frac{v_1}{r^3}, \, - \frac{v_2}{r^3} \right], \; r = \sqrt{v_1^2 + v_2^2}
\end{equation}

From now on we will be using $x, \, y, \, x', \, y'$, for being more straightforward than $v_1, v_2, v_3, v_4$.

The neural network we have constructed has six inputs: The coordinates of the second body: $x$ and $y$, the derivatives of the coordinates with respect to the independent variable t: $x'$, $y'$, the steplength $h$ and a "dummy" input with the constant value of 1. The input data consist of a matrix, where each column is a vector, intended to be used by the corresponding input of the network. The matrix satisfies the following equation
\begin{equation}
\label{input}
\left[x_n, \, y_n, \, x'_n, \, y'_n, \, h, \, 1 \right] = [\cos t_n, \, \sin t_n, \, -\sin t_n, \, \cos t_n, \, h, \, 1]
\end{equation}

With $t_{n+1} = t_n + h$, where $n = 1, 2, ..., N$ and $N$ is the number of total steps. Similarly, the target matrix satisfies the following equation
\begin{equation}
\label{target}
[x_{n+1}, \, y_{n+1}, \, x'_{n+1}, \, y'_{n+1}] = [\cos t_{n+1}, \, \sin t_{n+1}, \, -\sin t_{n+1}, \, \cos t_{n+1}]
\end{equation}

The input and the target data are generated through a certain procedure. In this procedure, we set a number of parameters, namely the integration interval $[t_s,t_e]$, the $x_0, y_0, x_0', y_0'$ values that correspond to ${\bf v_0}$ as notated in equation (\ref{secondorder}), and the constant steplength $h$. With the use of the integration interval and the steplength, the number of steps $N$ is calculated. For each step, the process generates the input and target data that correspond to the starting and ending point of the integration step respectively. Thus, two matrices are generated by using the theoretical solution of the problem and consist of $x, y, x', y'$ at the beginning and at the end of each integration step.

Apart from the input layer, the network comprises of thirteen additional layers, each consisting of a single neuron. Each layer is connected to one or more other layers, always in a feedforward fashion. Most of the connections have a fixed, unadjustable weight, whose value is equal to one. Practically, that means that the significance of the specific connections does not vary in the progress of the training phase. The connections that do have an adjustable weight are those related with the three coefficients of the Runge-Kutta method, namely $a_{21}$, $b_1$, and $b_2$. The coefficient $c_2$ is equal to $a_{21}$, therefore it does not need to be explicitly calculated through the use of the neural network.

\begin{figure}
\includegraphics[width = \textwidth]{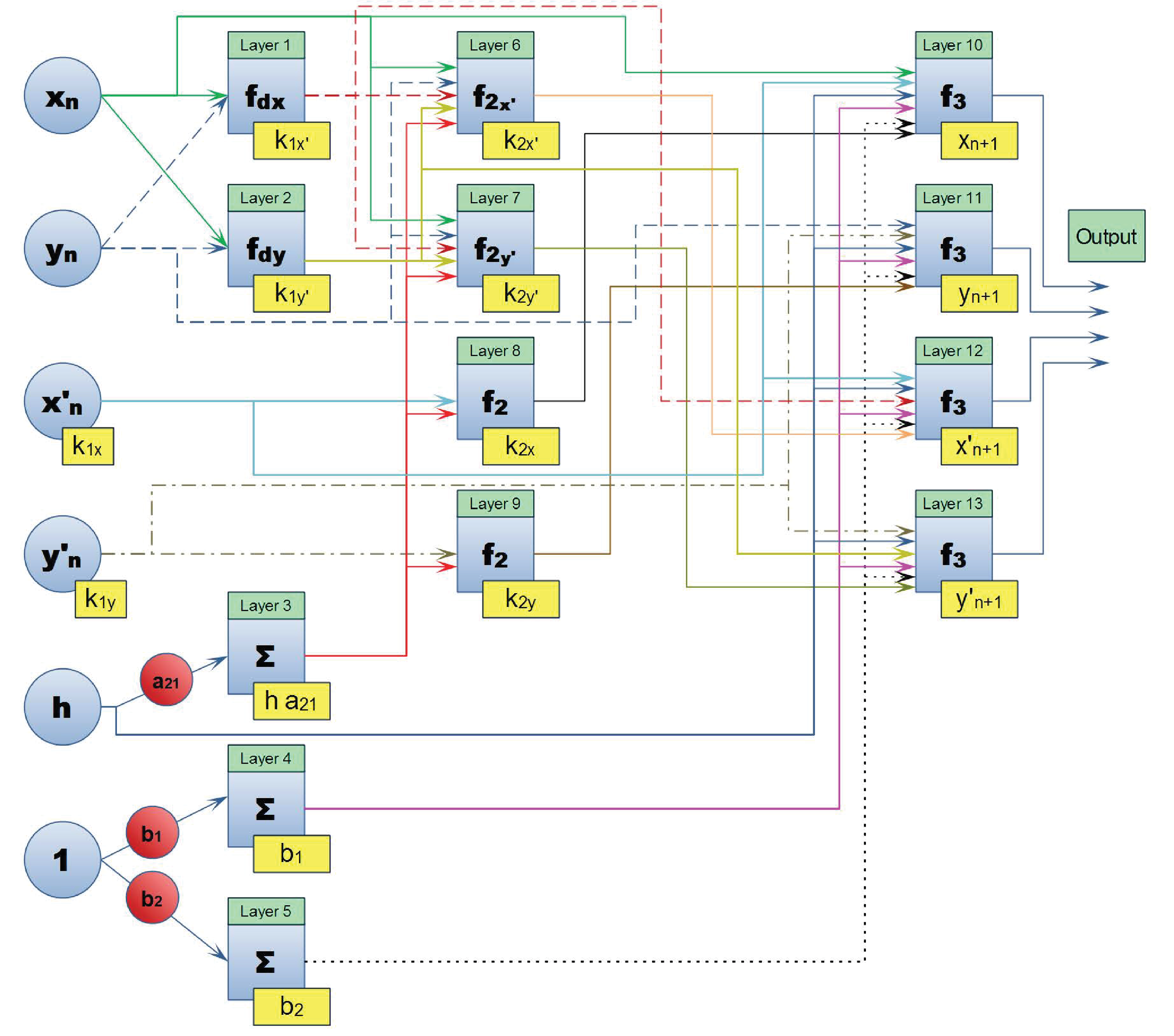}
\caption{The constructed neural network}
\end{figure}

The neural network is constructed in such manner as to replicate the actual Runge-Kutta methodology and produce the numerical solution for the two-body problem for each step of the integration. In contrast to the actual methodology, the approximate solution at the end of each step is not used as a starting point for the next step, but instead of this, theoretical values are used as starting points for each step's calculations.

Most of the layers do not use the standard summation function as the net input function, but a custom one, different in most of the cases. The activation function in every neuron is the identity function, described by the following equation
\begin{equation}
y = f(u) = u
\end{equation}
Therefore, the transfer function in each case is, in essence, the net input function. Below, there is the list of the inputs and the layers of the network, their input and output connections (with any of the 6 inputs or 13 layers) and in parentheses the expression that they represent:

Input 1:
\begin{itemize}
\item Output connections: layers 1 ($k_{1x'}$), 2 ($k_{1y'}$), 6 ($k_{2x'}$), 7 ($k_{2y'}$), 10 ($x_{n+1}$)
\item Value: $x_n$
\end{itemize}
Input 2:
\begin{itemize}
\item Output connections: layers 1 ($k_{1x'}$), 2 ($k_{1y'}$), 6 ($k_{2x'}$), 7 ($k_{2y'}$), 11 ($y_{n+1}$)
\item Value: $y_n$
\end{itemize}
Input 3:
\begin{itemize}
\item Output connections: layers 8 ($k_{2x}$), 10 ($x_{n+1}$), 12 ($x'_{n+1}$)
\item Value: $x'_n = k_{1x}$
\end{itemize}
Input 4:
\begin{itemize}
\item Output connections: layers 9 ($k_{2y}$), 11 ($y_{n+1}$), 13 ($y'_{n+1}$)
\item Value: $y'_n = k_{1y}$
\end{itemize}
Input 5:
\begin{itemize}
\item Output connections: layers 3 ($h \, a_{21}$), 10 ($x_{n+1}$), 11 ($y_{n+1}$), 12 ($x'_{n+1}$), 13 ($y'_{n+1}$)
\item Value: $h$
\end{itemize}
Input 6:
\begin{itemize}
\item Output connections: layers 4 ($b_1$), 5 ($b_2$)
\item Value: $1$
\end{itemize}
Layer 1:
\begin{itemize}
\item Input connections: inputs 1 ($x_n$), 2 ($y_n$)
\item Output connections: layers 6 ($k_{2x'}$), 7 ($k_{2y'}$), 12 ($x'_{n+1}$)
\item Net input function: $f_{x'}(x_n,y_n) = - \frac{x_n}{(\sqrt{x_n^2+y_n^2})^3}$
\item Output value: $k_{1x'}$
\end{itemize}
Layer 2:
\begin{itemize}
\item Input connections: inputs 1 ($x_n$), 2 ($y_n$)
\item Output connections: layers 6 ($k_{2x'}$), 7 ($k_{2y'}$), 13 ($y'_{n+1}$)
\item Net input function: $f_{y'}(x_n,y_n) = - \frac{y_n}{(\sqrt{x_n^2+y_n^2})^3}$
\item Output value: $k_{1y'}$
\end{itemize}
Layer 3:
\begin{itemize}
\item Input connections: input 5 ($h$)
\item Output connections: layers 6 ($k_{2x'}$), 7 ($k_{2y'}$), 8 ($k_{2x}$), 9 ($k_{2y}$)
\item Net input function: $\sum$
\item Output value: $h \, a_{21}$
\end{itemize}
Layer 4:
\begin{itemize}
\item Input connections: input 6 ($1$)
\item Output connections: layers 10 ($x_{n+1}$), 11 ($y_{n+1}$), 12 ($x'_{n+1}$), 13 ($y'_{n+1}$)
\item Net input function: $\sum$
\item Output value: $b_1$
\end{itemize}
Layer 5:
\begin{itemize}
\item Input connections: input 6 ($1$)
\item Output connections: layers 10 ($x_{n+1}$), 11 ($y_{n+1}$), 12 ($x'_{n+1}$), 13 ($y'_{n+1}$)
\item Net input function: $\sum$
\item Output value: $b_2$
\end{itemize}
Layer 6:
\begin{itemize}
\item Input connections: inputs 1 ($x_n$), 2 ($y_n$), layers 1 ($k_{1x'}$), 2 ($k_{1y'}$), 3 ($h \, a_{21}$)
\item Output connections: layer 12 ($x'_{n+1}$)
\item Net input function: $f_{2x'} = f_{x'}(x_n+h \, a_{21} \, k_{1x'}, \, y_n+h \, a_{21} \, k_{1y'})$
\item Output value: $k_{2x'}$
\end{itemize}
Layer 7:
\begin{itemize}
\item Input connections: inputs 1 ($x_n$), 2 ($y_n$), layers 1 ($k_{1x'}$), 2 ($k_{1y'}$), 3 ($h \, a_{21}$)
\item Output connections: layer 13 ($y'_{n+1}$)
\item Net input function: $f_{2y'} = f_{y'}(x_n+h \, a_{21} \, k_{1x'}, \, y_n+h \, a_{21} \, k_{1y'})$
\item Output value: $k_{2y'}$
\end{itemize}
Layer 8:
\begin{itemize}
\item Input connections: input 3 ($x'_n = k_{1x}$), layer 3 ($h \, a_{21}$)
\item Output connections: layer 10 ($x_{n+1}$)
\item Net input function: $f_{2} = x'_n + h \, a_{21} \, k_{1x} = x'_n \cdot (1+h \, a_{21})$
\item Output value: $k_{2x}$
\end{itemize}
Layer 9:
\begin{itemize}
\item Input connections: input 4 ($y'_n = k_{1y}$), layer 3 ($h \, a_{21}$)
\item Output connections: layer 11 ($y_{n+1}$)
\item Net input function: $f_{2} = y'_n + h \, a_{21} \, k_{1y} = y'_n \cdot (1+h \, a_{21})$
\item Output value: $k_{2y}$
\end{itemize}
Layer 10:
\begin{itemize}
\item Input connections: inputs 1 ($x_n$), 3 ($x'_n = k_{1x}$), 5 ($h$), layers 4 ($b_1$), 5 ($b_2$), 8 ($k_{2x}$)
\item Output connections: network output
\item Net input function: $f_{3} = x_n+h \cdot (b_1 \, k_{1x} + b_2 \, k_{2x})$
\item Output value: $x_{n+1}$
\end{itemize}
Layer 11:
\begin{itemize}
\item Input connections: inputs 2 ($y_n$), 4 ($y'_n = k_{1y}$), 5 ($h$), layers 4 ($b_1$), 5 ($b_2$), 9 ($k_{2y}$)
\item Output connections: network output
\item Net input function: $f_{3} = y_n+h \cdot (b_1 \, k_{1y} + b_2 \, k_{2y})$
\item Output value: $y_{n+1}$
\end{itemize}
Layer 12:
\begin{itemize}
\item Input connections: inputs 3 ($x'_n = k_{1x}$), 5 ($h$), layers 1 ($k_{1x'}$), 4 ($b_1$), 5 ($b_2$), 6 ($k_{2x'}$)
\item Output connections: network output
\item Net input function: $f_{3} = x'_n+h \cdot (b_1 \, k_{1x'} + b_2 \, k_{2x'})$
\item Output value: $x'_{n+1}$
\end{itemize}
Layer 13:
\begin{itemize}
\item Input connections: inputs 4 ($y'_n = k_{1y}$), 5 ($h$), layers 2 ($k_{1y'}$), 4 ($b_1$), 5 ($b_2$), 7 ($k_{2y'}$)
\item Output connections: network output
\item Net input function: $f_{3} = y'_n+h \cdot (b_1 \, k_{1y'} + b_2 \, k_{2y'})$
\item Output value: $y'_{n+1}$
\end{itemize}

\section{Method derivation}
\label{Sec_Derivation}

The derivation of the coefficients for the RK method described in (\ref{RK}), includes certain subroutines:
\begin{itemize}
\item Data generation
\item Neural network training
\item Conversion to fraction
\end{itemize}

During the data generation phase, we generate the input and target data that are going to be used to train the neural network. We have used various integration intervals $[t_s,t_e]$, from $[0,2\pi]$ up to $[0,200\pi]$ and various steplengths, from $h = \frac{\pi}{4}$ down to $\frac{\pi}{512}$. As we mentioned before, the input and the target data are described by the equations (\ref{input}) and (\ref{target}) respectively. Therefore, after the data generation, we obtain an input matrix of size $6 \times N$ and a target matrix of size $4 \times N$. Where $N$ is the number of steps and the following equation holds
\begin{equation}
N = \frac{t_e - t_s}{h}
\end{equation}

During the training phase, we use the generated data to train the neural network. The method used to conduct the training is a variation of the typical backpropagation method, called {\bf Gradient Descent with Momentum and Adaptive Learning Rate Backpropagation} or GDX. The back-propagation algorithm and its numerous variants constitute the most popular learning technique for feedforward neural networks \cite{haykin}. However, a shortcoming of the original back-propagation algorithm is that it can be easily trapped in local minima. In order to deal with this disadvantage, the addition of a momentum term has been proposed. GDX includes the momentum term in order to be able to escape from local minima, but also has been found to present faster convergence and lower training times compared to other competing methods \cite{yu}.

Instead of a standard error estimation function, we use a custom one for this purpose, which is shown below
\begin{equation}
||{\bf Output} - {\bf Target}||_{\infty} + |b_1 + b_2 - 1| + |a_{21} \, b_2 - \frac{1}{2}|
\end{equation}
The error to be minimized is the maximum absolute difference between the output and the target data, plus some added expressions to satisfy the algebraic conditions. The terms $|b_1 + b_2 - 1|$ and $|a_{21} \, b_2 - \frac{1}{2}|$ are used to satisfy the conditions (\ref{alg1}) and (\ref{alg2}) respectively. The coefficient $a_{21}$ is used interchangeably with $c_2$, due to (\ref{cons}).

As a result of the training phase, the obtained coefficients $a_{21}$, $b_1$ and $b_2$ satisfy the conditions to a certain degree of accuracy. Additionally, the Runge-Kutta method constructed with the generated coefficients can produce an output, sufficiently close to the target data.

During the last phase, the coefficient $a_{21}$ is converted into a fraction to simplify the method, with an insignificant loss of accuracy. The rest of the coefficients are calculated with the use of the fractional $a_{21}$, in order for the algebraic and consistency conditions, (\ref{alg1}), (\ref{alg2}) and (\ref{cons}) respectively, to be satisfied. The coefficients of the new method are presented in Table \ref{New_RK_table}.

For the current implementation, the number of maximum iterations was set at 10000. That value was selected empirically, as the neural network provided a solution after approximately 5000 iterations and effectively completed the training, as it could not improve on the performance any further.

\section{Results}
\label{Sec_Results}

The best method was provided by using the integration interval $[0,2\pi]$ and the steplength $h = \frac{\pi}{128}$. Training with other integration intervals and steplengths have returned similar results. $2\pi$ represents a full oscillation for the two-body problem, which in part explains why training over wider intervals does not yield better results. Apart from the new method, the training has also generated some well known classical methods, some of which are given later in this section.

We compare the new method with three classical two-stage explicit Runge-Kutta methods. The corresponding Butcher tableaus of all methods are shown in Tables \ref{New_RK_table}-\ref{Third_RK_table}.

\begin{table}[h!]
\begin{tabular}{l|l l}
 $\frac{11}{26}$ & $\frac{11}{26}$ & \\
 \hline
 & $-\frac{2}{11}$ & $\frac{13}{11}$
\end{tabular}
\caption{New method}
\label{New_RK_table}
\end{table}

\begin{table}[h!]
\begin{tabular}{l|l l}
 $1$ & $1$ & \\
 \hline
 & $\frac{1}{2}$ & $\frac{1}{2}$
\end{tabular}
\caption{Heun's method}
\end{table}

\begin{table}[h!]
\begin{tabular}{l|l l}
 $\frac{1}{2}$ & $\frac{1}{2}$ & \\
 \hline
 & $0$ & $1$
\end{tabular}
\caption{Midpoint method}
\end{table}

\begin{table}[h!]
\begin{tabular}{l|l l}
 $\frac{2}{3}$ & $\frac{2}{3}$ & \\
 \hline
 & $\frac{1}{4}$ & $\frac{3}{4}$
\end{tabular}
\caption{Third classical method}
\label{Third_RK_table}
\end{table}

\begin{figure}
\includegraphics[width = \textwidth]{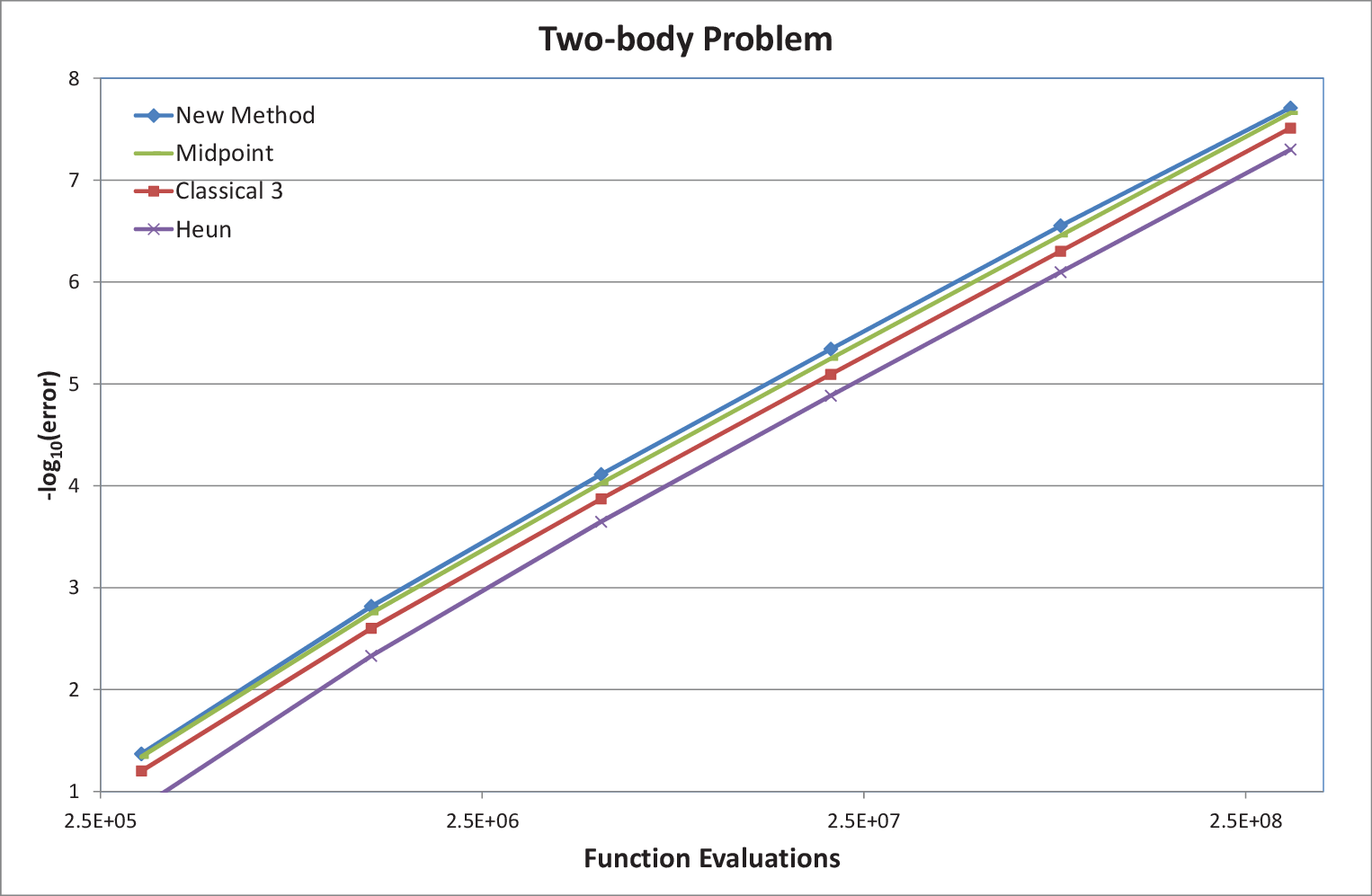}
\caption{Efficiency of the compared methods for the integration of the two-body problem}
\label{fig_comparison}
\end{figure}

A comparison between the classical methods and the new method is presented in figure \ref{fig_comparison}, which regards the numerical integration of the two-body problem over the interval $[0,1000]$. The vertical axis represents the accuracy as expressed by $-log_{10}$(maximum absolute error), while the horizontal axis represents the total number of function evaluations that are required for the computation. The latter is provided by the formula $F.E. = s \cdot N$, where $s=2$ and stands for the number of stages of the RK method and $N$ is the number of total steps. We can see that the new method is more efficient among all methods compared.

\section{Conclusions}
\label{Sec_Conclusions}

We have constructed an artificial neural network that can generate the optimal coefficients of an explicit two-stage Runge-Kutta method. The network is specifically designed to apply to the needs of the two-body problem, and therefore the resulting method is specialized in solving that problem. Following the implementation of the network, the latter is trained to obtain the optimal values for the coefficients of the method. The comparison has shown that the new method developed in this work is more efficient than other well known methods and thus proved the capability of the constructed neural network to create new efficient numerical algorithms.

\bibliographystyle{aipproc}

\end{document}